\pdfoutput=1

\documentclass[11pt,a4paper]{article}
\usepackage{emnlp2018_arxiv}
\usepackage{times}
\usepackage{latexsym}

\usepackage{url}

\usepackage{float}
\usepackage{placeins}
\usepackage{graphicx}
\usepackage{amsmath}
\usepackage{multirow}
\usepackage{tabularx}
\usepackage{pbox}
\usepackage{epstopdf}
\usepackage{epsfig}

\usepackage[TABTOPCAP]{subfigure} 

\usepackage{enumitem}
\setlist{noitemsep}
\aclfinalcopy 

\title{Rapid Customization for Event Extraction}

\author{
 Yee Seng Chan, Joshua Fasching, Haoling Qiu, \and Bonan Min\\
  Raytheon BBN Technologies\\
  {\tt \{yeeseng.chan,joshua.fasching\}@raytheon.com}\\
  {\tt \{haoling.qiu,bonan.min\}@raytheon.com}
}

\date{}

\begin{document}
\maketitle
\begin{abstract}
  We present a system for rapidly customizing event extraction capability
  to find new event types and their arguments.
  The system allows a user to find, expand and filter event triggers
  for a new event type by exploring an unannotated corpus. The system will then
  automatically generate mention-level event annotation automatically,
  and train a Neural Network model for finding the corresponding event.
  Additionally, the system uses the ACE corpus to train an argument
  model for extracting Actor, Place, and Time arguments for any event types, including ones not seen in its training data.
  Experiments show that with less than 10 minutes of human effort per event type,
  the system achieves good performance for 67 novel event types.
  The code, documentation, and a demonstration video will be released as open source on {\url github.com}.
\end{abstract}

\section{Introduction}

\begin{figure*}[t]
\begin{center}
\scalebox{1.0}{%
\includegraphics[width=1.0\linewidth]{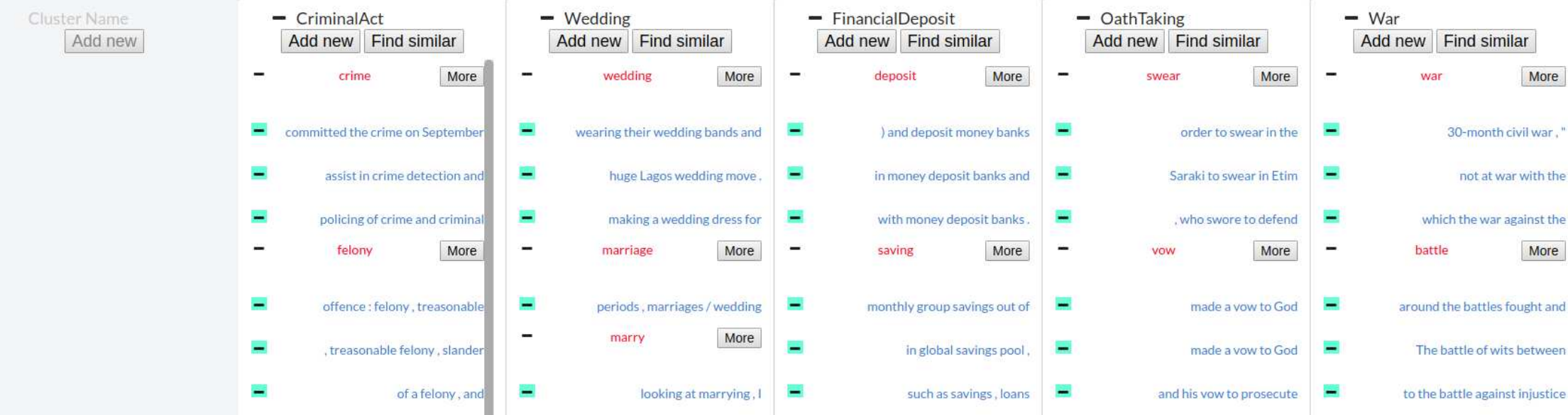}}
\end{center}
\vspace{-0.3cm}
\caption{A user interface that allows a user to provide, expand, and filter event triggers for new types. Each column shows an event type name at the top, followed by event triggers (in red) and text snippets (clickable to expand to full sentence) mentioning these triggers. The user can drag and drop a trigger or sentence from one event to another. The user can also click on ``$-$'' to remove an event, a trigger with its snippets, or just a snippet. The first column to the left is reserved for displaying similar triggers found by word similarity tools.}
\label{fig2}
\end{figure*}

Event extraction is the task of identifying events of interest with associated participating arguments in text.
For instance, given the following sentence:

{\bf S1}: 21 people were wounded in Tuesday's southern Philippines airport blast.

We want to recognize that there are two events (Injury and Attack),
anchored or {\it triggered} by the words ``wounded'' and ``blast'' respectively.
We also recognize that ``21 people'' and ``airport'' are the participating arguments of both events,
taking the event argument roles affected-Actor and Place respectively.
Event extraction is important for understanding complex scenarios, intelligence analysis, etc.

Current supervised event extraction systems usually assume a predefined event ontology
and learn from a corpus of manually labeled examples that are specific to that ontology.
For instance, the popular Automatic Content Extraction (ACE)~\cite{conf-lrec-DoddingtonMPRSW04} corpus contains
more than 500 documents manually annotated with examples for 33 event types such as Attack, Injury, Election, etc.
To extend extraction capabilities to a new event type, one needs to
manually annotate a large amount of training examples for that new event type. This is very labor intensive.

In this paper, we present a system that uses a minimally supervised approach for rapid extension of
extraction capabilities to a large number of novel event types.

Briefly, we first manually provide a few triggers or keywords for each novel event type.
We then automatically expand these initial triggers by leveraging
WordNet~\cite{Miller:1995:WLD:219717.219748} and word embeddings \cite{Baroni2014}.
These expanded trigger words are then adjudicated, culminating in a final set of trigger words.
Our system then automatically gathers event trigger training examples via distant supervision
\cite{mintz-EtAl:2009:ACLIJCNLP,Craven:1999:CBK:645634.663209} on an unannotated corpus.
We summarize the contributions of this paper as follows:
\begin{itemize}
\item We present an approach to rapidly gather event trigger examples for $new$ event types, with minimal human effort.
\item We show how to leverage annotations of $existing$ event types, to train a classifier that extracts
event arguments for the $new$ event types.
\item We demonstrate the practical utility of our approach by applying it on a set of 67 novel event types.
\item We develop a User Interface (UI) to further expedite and improve the time efficiency of our approach.
The code, documentation, and a demonstration video will be released as open source on {\url github.com}.
\end{itemize}

We first describe the task of event extraction in the next section. We then show how to rapidly gather
event trigger examples for new event types in Section \ref{section:trigger}.
In Section \ref{section:argument}, we show how to extract event arguments for novel event types.
In Section \ref{section:model}, we describe our model architecture.
We present experiment results in Section \ref{section:experiment} and perform analysis in Section \ref{section:analysis}.
We discuss related work in Section \ref{section:related} before concluding in Section \ref{section:conclusion}.



\section{Event Extraction}

Given an English sentence, we perform event extraction using a two-stage process:
\begin{itemize}
\item Trigger classification: Labeling words in the sentence with their predicted event type (if any).
For instance, in sentence S1, the extraction system should label ``wounded'' as a trigger of an
Injury event, and label ``blast'' as a trigger of an Attack event.
\item Argument classification: If a sentence contains predicted event triggers $\{t_{i}\}$,
we pair each $t_i$ with each entity and time mention \begin{math}\{m_j\}\end{math} in the sentence to generate
candidate event arguments. Given a candidate event argument $(t_{i},m_{j})$, the system
predicts its associated event role (if any).
For instance, given the candidate event argument (``wounded'', ``airport''), the system should predict
the event role $Place$.
\end{itemize}

\section{Trigger Examples for New Events}\label{section:trigger}

In this section, we describe how we rapidly gather event trigger examples for new event types
with minimal human effort, aided by the UI shown in Figure \ref{fig2}. Our approach is as follows:

\paragraph{Initial triggers} Given a new event type, we first ask a user to
provide a few discriminative keywords $\mathcal{K}$.

\paragraph{Expanded triggers} Next, we add words that are related to $\mathcal{K}$ using two approaches:
\begin{itemize}
\item Leverage word embeddings to gather words $\mathcal{W}_{e}$ that have a high cosine similarity to $\mathcal{K}$.
\item Leverage WordNet to gather hyponyms\footnote{X is a hyponym of Y if X is a (kind of) Y.}
$\mathcal{W}_{h}$ of $\mathcal{K}$.
\end{itemize}
We then ask the user to filter $\mathcal{W}_{e}$ and $\mathcal{W}_{h}$,
only keeping words $\mathcal{W}_{r} \subseteq (\mathcal{W}_{e} \cup \mathcal{W}_{h})$  that are relevant to the new event type.

\paragraph{Distant supervision} We then locate occurrences of $\mathcal{K} \cup \mathcal{W}_{r}$
in an unannotated text corpus and gather them as event trigger examples for the new event type.

In practice, over a set of 67 new event types, the user spent an average of 4.5 minutes to
provide 8.6 initial triggers, and 5 minutes interacting with the UI to filter the expanded triggers, 
for a total of less than 10 minutes per event type.

\section{Argument Extraction for New Events}\label{section:argument}

\begin{figure*}[t]
\begin{center}
\scalebox{0.6}{%
\includegraphics[width=1.0\linewidth]{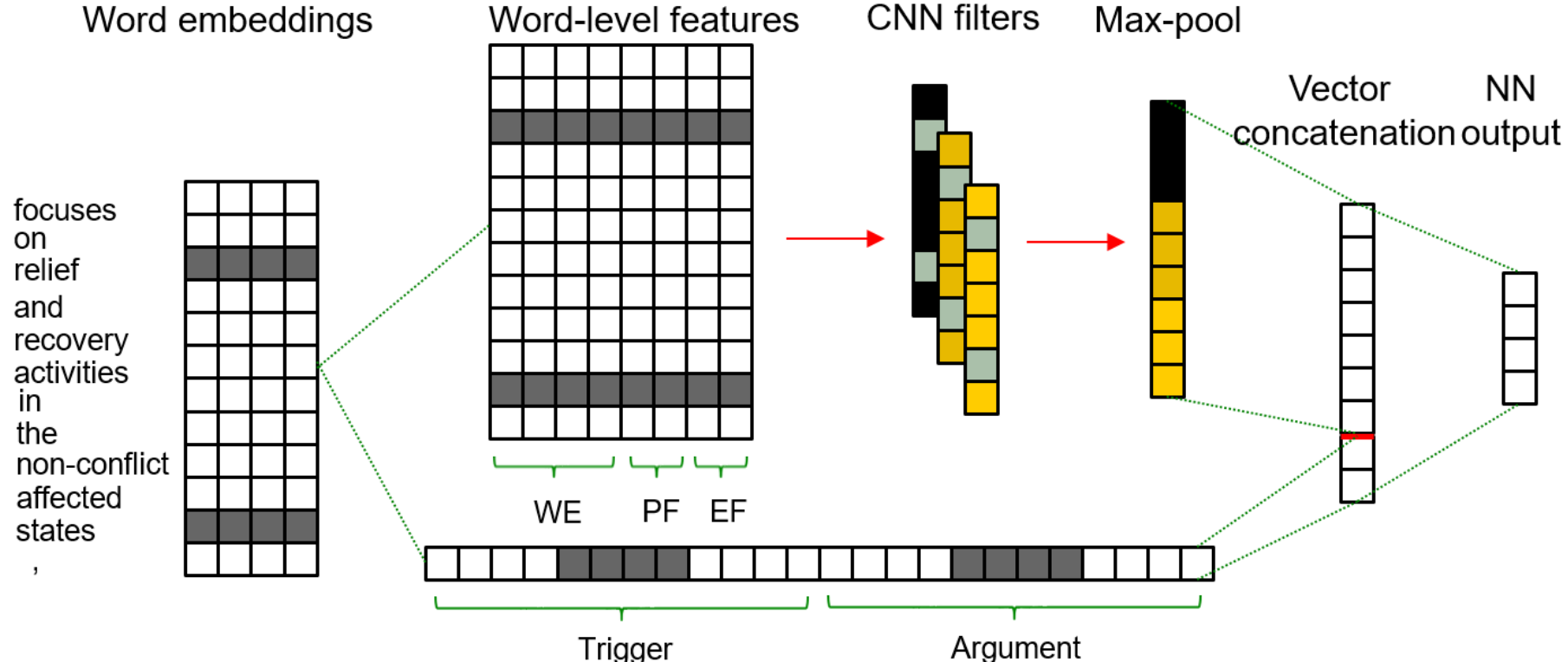}}\label{fig:cnn}
\end{center}
\vspace{-0.3cm}
\caption{A Neural Network model for event argument classification. 
The model for trigger classification omits the argument features that are shown at the bottom.
Chen et al. \shortcite{chen-EtAl:2015:ACL-IJCNLP2} models event type using event embeddings features (EF),
but we omit this when training a {\it generic} event argument model.}
\label{fig4}
\end{figure*}

Current event argument extraction systems
learn from training examples annotated for a predefined ontology.
For instance, the ACE corpus contains argument annotations such as the Agent and Victim of
Injury events, the Attacker and Target of Attack events, etc.

When given a candidate event argument $(t_i,m_j)$ where $t_i$ is a predicted event trigger,
state-of-the-art systems such as DMCNN \cite{chen-EtAl:2015:ACL-IJCNLP2},
usually use the predicted event type of $t_i$ as a feature in its model.
While this generally helps to improve performance, the learned system will not be able
to label arguments for new event types not seen in the training data.

In this paper, we propose a simple approach to learn a $generic$ event argument model that
could be applied to extract Actor, Place, and Time arguments for new event types.
In particular, we map Actor-$like$ argument roles (e.g. Attacker, Victim, Target, etc.)
in ACE to a common Actor role label, and omit using event type of $t_i$ as a feature
in the argument role classifier. The Place and Time arguments in ACE are kept as they are.
The complete list of ACE event argument roles that we mapped to Actor are as follows:
\begin{itemize}
\item Person, Agent, Victim, Artifact, Buyer, Seller, Giver, Recipient, Org, Attacker, Target,
Entity, Defendant, Prosecutor, Plaintiff
\item Adjudicator of event types: Convict, Sentence, Fine, Acquit, Pardon
\end{itemize}

Using the above approach, we build a classifier that could extract Actor, Place, and Time arguments
for any event type.

\section{Neural Network Model}\label{section:model}

To perform event trigger and argument classification, we developed a convolution neural network (CNN)
model based on the work of Chen et al. \shortcite{chen-EtAl:2015:ACL-IJCNLP2},
which achieves competitive performance for event extraction.
The authors also added a dynamic multi-pooling (DM) layer for CNN, which provides further improvements
for event extraction over CNN. Since our focus in this paper is on rapid customization of extraction capabilities
to new event types, we leave the implementation of DM to future work.

We show our CNN model in Figure \ref{fig:cnn}, which primarily uses unsupervised pre-trained word embeddings
as input features.
In this work, we use the word embeddings\footnote{http://clic.cimec.unitn.it/composes/semantic-vectors.html}
trained by Baroni et al. \shortcite{Baroni2014}, which achieved state-of-the-art results in a variety of NLP tasks.
The embeddings provide representation for about 300K words (lower-cased, non-lemmatized).
It was trained on the concatenation of ukWaC\footnote{http://wacky.sslmit.unibo.it}, the English Wikipedia,
and the British National Corpus
using the Skip-gram model with 5-word context window, 10 negative samples, and 400 dimensions.
Briefly, the Skip-gram model trains the embeddings of words $w_{1}, w_{2}, ..., w_{m}$ by maximizing
the average log probability:
\begin{equation} \label{eq:skipgram1}
\frac{1}{m}\sum_{t=1}^{m}\sum_{-c \le j \le c, j \ne 0} \text{log} p(w_{t+j}|w_{t})
\end{equation}
where $c$ is the window size and $p(w_{t+j}|w_{t})$ is defined as:
\begin{equation} \label{eq:skipgram2}
\frac{\text{exp}(e_{t+j}^{'T}e_{t})}{\sum_{w=1}^{m}\text{exp}(e_{w}^{'T}e_{t})}
\end{equation}
where $m$ is the vocabulary of the unlabeled text and $e_{i}^{'}$ is another embedding for $e_{i}$

Given a sentence, our CNN model constructs word-level features using the following:
\begin{itemize}
\item Word Embeddings (WE): We represent each word in the sentence with its word embeddings vector.
\item Position Feature (PF): For every word $w_{i}$ in the sentence, this captures its
relative token distance to the predicted trigger or candidate argument. For instance, in Figure \ref{fig:cnn},
the relative distances of ``on'' and ``recovery'' to the trigger ``relief'' are $-1$ and $+2$ respectively.
These distance values to the trigger are represented by $PF_{trigger}$ embedding vectors which is
learnt during training. When performing argument classification, 
we similarly use $PF_{arg}$ embedding vectors to capture token distances to the candidate argument.
\end{itemize}

The WE and PF embeddings are fed into the convolution and max-pooling layers.
As shown at the bottom of Figure \ref{fig:cnn}, we also construct a vector consisting of
the word embeddings of the candidate trigger, the candidate argument,
and the embeddings of words in their local window (i.e. words on their left and right).
This vector is then concatenated with the max-pooled vector and fed into the output layer for classification.
In this paper, we set the dimension of PF to 5, and the window size as 3.
For more details, we refer the reader to \cite{chen-EtAl:2015:ACL-IJCNLP2}.

\section{Experiments}\label{section:experiment}

We use the following criteria to judge the correctness of our event extractions:
\begin{itemize}
\item A trigger is correctly classified if its event subtype and offsets match those of a reference trigger.
\item An argument is correct classified if its event subtype, event argument role, and offsets match any
of the reference event arguments
\end{itemize}

For all our experiments reported in this paper, we follow \cite{chen-EtAl:2015:ACL-IJCNLP2}
by tuning the model parameters (batch size and number of CNN filter maps) on the development documents.
We also tried different number of epoches: 5, 10, or 20. In addition, since the ratio of positive vs negative examples
is relatively skewed (for instance, most words in a sentence are not triggers), we also tried different
weights for the positive examples: 1, 3, 5, or 10.
Finally, we follow \cite{chen-EtAl:2015:ACL-IJCNLP2} by using the Adadelta update rule with
parameters $\rho=0.95$ and $\epsilon=1e^{-6}$, and a dropout rate of 0.5.


To evaluate the effectiveness of our event extraction system in customizing extractors
for new event types, we present experiment results in the rest of this section,
based on the Common Core Ontologies\footnote{https://github.com/CommonCoreOntology} (CCO).
CCO comprises 11 ontologies and is aimed at representing semantics for many domains of interests.
The CCO event ontology covers a wide variety of actions, events, processes, and change of states.
We sampled 67 event types that are not in existing event schemas
(such as ACE and TAC-KBP\footnote{https://tac.nist.gov/2017/KBP/Event/index.html}),
to evaluate how well our system does on novel event types.
As our experiment corpus $\mathcal{C}$, we use 6,000 allafrica.com news articles, published between 2016-2017.

\subsection{Trigger Classification}\label{section:experiment_trigger}

\begin{table}[t]
\centering
\begin{tabular}{l|c|c|c|c|c}
\hline & $\mathcal{C}_{d}$ & $\mathcal{C}_{h}$ & $\mathcal{C}_{s}$ & Dev & Test\\
\hline \#Docs & 818 & 618 & 618 & 274 & 273\\
\#Egs & 1674 & 1171 & 1258 & 643 & 752 \\
\hline
\end{tabular}
\caption{A set of 1,365 allafrica.com articles are gathered via distance supervision (DS) of trigger examples.
A set of 274 and 273 documents are manually annotated and set aside as dev and test data.
The DS examples within the remaining set of 818 documents are also manually filtered for correctness,
resulting in a smaller set of adjudicated examples associated with 618 documents $\mathcal{C}_{h}$.}
\label{table:data}
\end{table}

\begin{table}[t]
\centering
\begin{tabular}{l|c|c|c}
\hline & Precision & Recall & F1\\
\hline $T_{d}$ & 0.69 & 0.50 & 0.58\\
$T_{h}$ & 0.69 & 0.46 & 0.55\\
$T_{s}$ & 0.62 & 0.40 & 0.48\\
\hline
\end{tabular}
\caption{Event trigger results on new event types.}
\label{table:trigger_test}
\end{table}

Given the set of 67 new event types, we first ask our annotator to provide an initial set
of keywords or triggers for each event type, then perform expansion and filtering as described in
Section \ref{section:trigger}. On average for each event type, we obtain a set of 21.5 final set of triggers
from an initial set of 8.6 triggers.

Applying the distant supervision (DS) approach, we use the final set of triggers
to automatically gather trigger examples from $\mathcal{C}$ as follows:
\begin{itemize}
\item Automatically gather a maximum of 60 DS trigger examples per event type. When doing this,
we gathered trigger examples from a total of 1,365 documents.
\item We randomly do a 60/20/20 split of these documents into training, development, and test data.
For instance, as shown in Table \ref{table:data}, we obtain a set of 818 documents $\mathcal{C}_{d}$,
which we use to train a DS trigger model $T_{d}$.
\item $\mathcal{C}_{d}$ contains only DS trigger examples. To compare against the traditional approach 
of learning from manually annotated examples, we adjudicated and filtered $\mathcal{C}_{d}$,
obtaining a smaller set of adjudicated examples associated with 618 documents $\mathcal{C}_{h}$, 
which we use to train a supervised trigger model $T_{h}$.
\end{itemize}

We tune model parameters for $T_{d}$ and $T_{h}$ on the development data and report their performance
on the test data in Table \ref{table:trigger_test}. The results show that our proposed
DS trigger model $T_{d}$ achieves an F1 score of 0.58 on the test data,
while the traditional approach of training a supervised trigger model $T_{h}$ achieves an F1 score of 0.55.
We were expecting the manually supervised model $T_{h}$ to perform better than the DS model $T_{d}$.
One possible explanation for the observed lower performance of $T_{h}$ is that
$\mathcal{C}_{h}$ is substantially smaller than $\mathcal{C}_{d}$. 
We will show this is exactly the case in Section \ref{section:analysis}.

\subsection{Argument Classification}\label{section:experiment_argument}

\begin{table}[t]
\centering
\begin{small}
\begin{tabular}{l|c c c|c c c}
\hline Model & \multicolumn{3}{c|}{Overall} & \multicolumn{3}{c}{F1}\\
 & P & R & F1 & Actor & Place & Time\\
\hline $A_{s}$ & 0.53 & 0.47 & 0.50 & 0.47 & 0.43 & 0.64 \\
$A_{m}$ & 0.65 & 0.41 & 0.50 & 0.49 & 0.37 & 0.61 \\
$A_{n}$ & 0.41 & 0.62 & 0.49 & 0.49 & 0.45 & 0.62 \\
\hline
\end{tabular}
\end{small}
\caption{Event argument results using gold triggers. 
$A_{s}$: argument model trained as normal, but predictions and reference roles are mapped to a common Actor role.
$A_{m}$: argument model by mapping Actor-$like$ argument roles to a common Actor role.
$A_{n}$: Since the motivation is for the argument model to decode on new event types not previously seen during training, we conduct $n$-fold experiments, where in each fold $i$, we omit (Actor, Place, Time) argument examples associated with event type $i$ from the training data.
}
\label{table:argument_gold}
\end{table}

\begin{table}[t]
\centering
\begin{small}
\begin{tabular}{l|c c c|c c c}
\hline Model & \multicolumn{3}{c|}{Overall} & \multicolumn{3}{c}{F1}\\
 & P & R & F1 & Actor & Place & Time\\
\hline $A_{s}$ & 0.38 & 0.34 & 0.36 & 0.34 & 0.31 & 0.45 \\
$A_{m}$ & 0.46 & 0.28 & 0.35 & 0.34 & 0.27 & 0.43 \\
$A_{n}$ & 0.32 & 0.44 & 0.37 & 0.36 & 0.32 & 0.53 \\
\hline
\end{tabular}
\end{small}
\caption{Event argument results using predicted triggers.}
\label{table:argument_predict}
\end{table}

In Section \ref{section:argument}, we describe our approach of learning an event argument model that
could be applied to extract Actor, Place, and Time arguments for new event types, by
mapping Actor-$like$ argument roles in ACE to a common Actor role. In this section, we evaluate this
approach by performing experiments on the ACE corpus.

We applied the above mapping approach to learn an argument model $A_{m}$ on the ACE training data,
showing its performance on the ACE test data
when using gold triggers (Table \ref{table:argument_gold}) and predicted triggers (Table \ref{table:argument_predict}).
In the second column of the tables, we show the overall Precision (P), Recall (R), and F1 scores
over Actor, Place, and Time in aggregate. Next, we also show the F1 scores for each individual role.

For comparison, we evaluated an alternative approach where we train an event argument model
as per normal (i.e. using the original ACE event roles), but report results after mapping predictions
and reference roles to a common Actor role. We show the results of this experiment as rows $A_{s}$.

We note that both $A_{s}$ and $A_{m}$ train on the entire ACE training data. However, the motivation
of our approach is for the argument model to decode on new event types not previously seen in its training data.
Hence, we conduct an additional set of leave-1-out experiments $A_{n}$. Specifically, ACE defines
event types at a coarse-grained (8 types) level and a fine-grained (33 types) level.
For instance, the $Attack$ and $Demonstrate$ event types are grouped into $Conflict$ as a coarser-grained event type.
We thus perform an 8-fold experiment where in each fold $i$, we omit from the training data,
(Actor, Place, Time) argument examples associated with event type $i$.
We then proceed to calculate test performance for that fold, on just the
argument examples associated with event type $i$, which was not seen during training. We aggregate the
test results over the 8 folds and present them as row $A_{n}$. We note that $A_{n}$ achieves reasonable 
performance when compared against $A_{s}$ and $A_{m}$. In particular, we note from Table \ref{table:argument_gold}, 
which shows argument extraction performance when using gold triggers (so that we could isolate the performance of the argument extractor),
that the overall F1 score of $A_{n}$ is very close to that of $A_{m}$.

\section{Analysis}\label{section:analysis}

In Section \ref{section:experiment_trigger}, we note that, contrary to our expectations,
$T_{h}$ has a lower performance than $T_{d}$. $\mathcal{C}_{d}$ (used to train $T_{d}$)
consists entirely of trigger examples gathered
via distance supervision. $\mathcal{C}_{h}$ (used to train $T_{h}$) is a subset of $\mathcal{C}_{d}$,
where only trigger examples that are judged to be correct for the event types are kept.
Table \ref{table:data} shows that $\mathcal{C}_{h}$ contains substantially fewer examples than $\mathcal{C}_{d}$
(1171 examples in 618 documents vs 1674 examples in 818 documents). To investigate the difference
in number of examples as a plausible reason for $T_{d}$'s higher performance,
we randomly down-sampled $\mathcal{C}_{d}$ to have the same number of documents as $\mathcal{C}_{h}$.
From the resulting $\mathcal{C}_{s}$ (which still consists entirely of distant supervision examples),
we trained the trigger model $T_{s}$ and show its results in Table \ref{table:trigger_test}. Here, we see that
$T_{s}$ performs worse than $T_{h}$, which matches our expectations.

In Section \ref{section:experiment_argument}, we provide experiment results of our argument model on the ACE corpus.
We now evaluate its performance on the 273 test documents listed in Table \ref{table:data},
which contains 67 new event types not found in ACE.
Using $T_{h}$ and $A_{m}$ as the trigger and argument models on these documents,
we randomly selected 78 Actor, 8 Place, and 14 Time arguments (for a total of 100)
predicted by $A_{m}$. Of these, we determine that 62 Actor, 7 Place, and 10 Time arguments are correctly predicted,
for an overall precision of 0.79.

\section{Related Work}\label{section:related}

Event extraction are often sentence-level pipeline~\cite{ahn:2006:ARTE} models with high-level features~\cite{Huang:2012:MTC:2900929.2900964, ji-grishman:2008:ACLMain}. There are growing interests in Neural Network models~\cite{chen-EtAl:2015:ACL-IJCNLP2} for event extracion. Chen et al., \shortcite{chen-EtAl:2015:ACL-IJCNLP2} apply dynamic multi-pooling CNNs for event extraction in a pipelined framework. ~\cite{nguyen-cho-grishman:2016:N16-1} propose joint event extraction using recurrent neural networks.

Nguyen et al. \shortcite{nguyen-EtAl:2016:RepL4NLP} proposed a two-stage Neural Network model for event type extension. Given a new event type with a small set of seed examples, they leverage examples from other event types. In another work, 
Peng et al. \shortcite{PengSoRo16} developed a minimally supervised approach to event trigger extraction by leveraging 
trigger examples gathered from the ACE annotation guidelines.

A closely related direction is rapid customization~\cite{gupta:vizacl14,akbik-konomi-melnikov:2013:SystemDemo} of IE systems. The ICE (Integrated Customization Environment) system~\cite{he-grishman:2015:demos} allows a user to create an information extraction (IE) system for a new domain by building new classes of entities and relations interactively. The main ideas are user-in-the-loop entity set expansion and boostrap learning for relation extraction. The WIZIE~\cite{li-EtAl:2012:Demo} system guides users to write rules for IE.

\section{Conclusion and Future Work}\label{section:conclusion}

We presented a system which allows a user to rapidly build event extractors
 to find new types of events and their arguments. We plan to use clustering
 techniques to automatically discover salient event trigger words in a new
corpus, to further reduce human effort involved in the customization process.

\section*{Acknowledgments}
This work was supported by DARPA/I2O Contract
No. W911NF-18-C-0003 under the World Model-
ers program. The views, opinions, and/or findings
contained in this article are those of the author and
should not be interpreted as representing the offi-
cial views or policies, either expressed or implied,
of the Department of Defense or the U.S. Govern-
ment. This document does not contain technology
or technical data controlled under either the U.S.
International Traffic in Arms Regulations or the
U.S. Export Administration Regulations.


\bibliography{emnlp2018}
\bibliographystyle{acl_natbib_nourl}

\end{document}